\documentclass[11pt,a4paper]{article}
\usepackage[hyperref]{acl2020}
\usepackage{times}
\usepackage{latexsym}
\usepackage{hyperref}
\usepackage{url}
\usepackage{graphicx}
\usepackage{amsmath}
\usepackage{bm, upgreek}
\usepackage{amssymb}
\usepackage{array, booktabs}
\usepackage{tabularx}
\usepackage{multirow}
\usepackage{mathtools}
\usepackage{arydshln}
\usepackage{tabularx}
\usepackage{color}
\usepackage{siunitx}
\usepackage{wrapfig}
\usepackage{changepage}
\usepackage{subcaption}
\usepackage{threeparttable}
\usepackage{boldline}
\usepackage{blindtext}
\usepackage{stfloats}

\usepackage{microtype}

\aclfinalcopy 
\def\aclpaperid{1800} 

\newcommand{\ours}{\textsc{Sparc}}
\newcommand{\fasterleast}{45}

\newcommand{\squadopen}{SQuAD-Open}
\newcommand{\trec}{CuratedTREC}
\usepackage{comment}

\newcommand{\am}{{\bm a}}

\newcommand{\xm}{{\bm x}}
\newcommand{\qm}{{\bm q}}

\newcommand{\hb}{\mathbf{h}}
\newcommand{\sbo}{\mathbf{s}}

\newcommand{\Hb}{\mathbf{H}}

\newcommand{\Sb}{\mathbf{S}}

\newcommand{\argmax}{\textrm{argmax}}

\title{Contextualized Sparse Representations for\\Real-Time Open-Domain Question Answering}

\author{Jinhyuk Lee$^{1}$ \quad Minjoon Seo$^{2, 3}$ \quad Hannaneh Hajishirzi$^{2, 4}$ \quad Jaewoo Kang$^{1}$\\
Korea University$^{1}$ \quad University of Washington$^{2}$ \\ Clova AI, NAVER$^{3}$ \quad Allen Institute for AI$^4$\\
\texttt{\{jinhyuk\_lee,kangj\}@korea.ac.kr}\\  \texttt{\{minjoon,hannaneh\}@cs.washington.edu}\\
}

\date{}

\begin{document}
\maketitle
\begin{abstract}
Open-domain question answering can be formulated as a phrase retrieval problem, in which we can expect huge scalability and speed benefit but often suffer from low accuracy due to the limitation of existing phrase representation models.
In this paper, we aim to improve the quality of each phrase embedding by augmenting it with a contextualized sparse representation (\ours).
Unlike previous sparse vectors that are term-frequency-based (e.g., tf-idf) or directly learned (only few thousand dimensions), we leverage rectified self-attention to indirectly learn sparse vectors in n-gram vocabulary space.
By augmenting the previous phrase retrieval model~\cite{seo2019real} with \ours, we show 4\%+ improvement in CuratedTREC and SQuAD-Open.
Our CuratedTREC score is even better than the best known \emph{retrieve \& read} model with at least \fasterleast x faster inference speed.\footnote{Code available at \href{https://github.com/jhyuklee/sparc}{https://github.com/jhyuklee/sparc}.}

\end{abstract}

\section{Introduction}
Open-domain question answering (QA) is the task of answering generic factoid questions by looking up a large knowledge source, typically unstructured text corpora such as Wikipedia, and finding the answer text segment~\cite{chen2017reading}.
One widely adopted strategy to handle such large corpus is to use an efficient document (or paragraph) retrieval technique to obtain a few relevant documents, and then use an accurate (yet expensive) QA model to \emph{read} the retrieved documents and find the answer~\citep{chen2017reading,wang2018r,das2018multistep,yang2019end}.

More recently, an alternative approach formulates the task as an end-to-end phrase retrieval problem by encoding and indexing every possible text span in a dense vector offline~\cite{seo2018phrase}.
The approach promises a massive speed advantage with several orders of magnitude lower time complexity, but it performs poorly on entity-centric questions, often unable to disambiguate similar but different entities such as ``\textit{1991}'' and ``\textit{2001}'' in dense vector space.
To alleviate this issue,~\citet{seo2019real} concatenate a term-frequency-based sparse vector with the dense vector to capture lexical information. However, such sparse vector is identical across the document (or paragraph), which means every word's importance is equally considered regardless of its context (Figure~\ref{fig:overview}). 

\begin{figure}[t]
\begin{center}
\includegraphics[height=5.7cm]{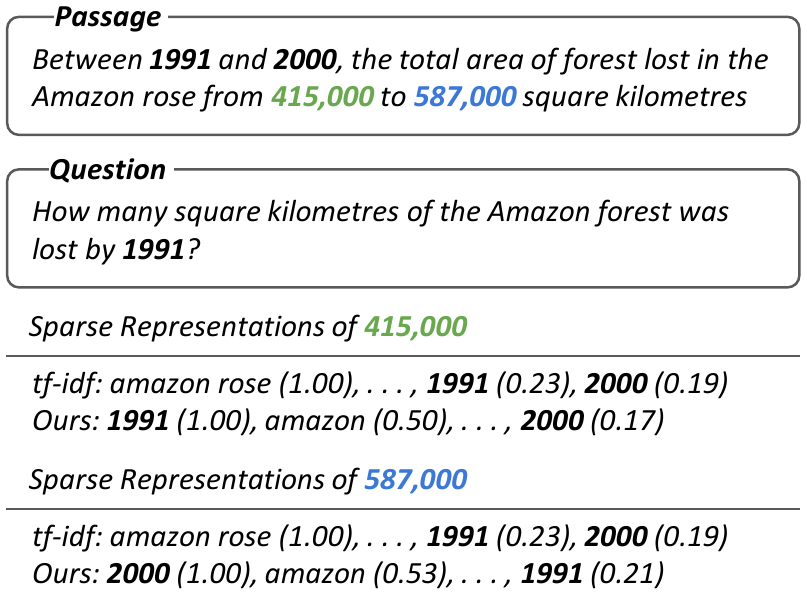}
\end{center}\vspace{-.25cm}
\caption{An example of sparse vectors given a context from SQuAD. While \textit{tf-idf} has high weights on infrequent n-grams, our contextualized sparse representation (\ours) focuses on sematically related n-grams.}\vspace{-.3cm}
\end{figure}\label{fig:overview}

In this paper, we introduce a method to learn a Contextualized Sparse Representation (\ours) for each phrase and show its effectiveness in open-domain QA under phrase retrieval setup.
Related previous work (for a different task) often directly maps dense vectors to a sparse vector space~\citep{faruqui2015sparse,subramanian2018spine}, which can be at most only a few thousand dimensions due to computational cost and small gradients.
We instead leverage rectified self-attention weights on the neighboring n-grams to scale up its cardinality to n-gram vocabulary space (billions), allowing us to encode rich lexical information in each sparse vector.
We \emph{kernelize}\footnote{Our method is inspired by the kernel method in SVMs~\citep{cortes1995support}.} the inner product space during training to avoid explicit mapping and obtain memory- and computational efficiency.

~\ours\ improves the previous phrase retrieval model, DenSPI~\cite{seo2019real} (by augmenting its phrase embedding), by more than 4\% in both \trec~and \squadopen.
In fact, our \trec~result achieves the new state of the art even when compared to previous retrieve \& read approaches, with at least \fasterleast x faster speed.

\section{Background}
We focus on open-domain QA on unstructured text where the answer is a text span in a textual corpus (e.g., Wikipedia). 
Formally, given a set of $K$ documents $\xm^1, \dots, \xm^K$ and a question $\qm$, the task is to design a model that obtains the answer $\hat{\am}$ by 
$\hat{\am} = \argmax_{\xm_{i:j}^k} F(\xm_{i:j}^k, \qm), $ where $F$ is the score model to learn and $\xm_{i:j}^k$ is a phrase consisting of words from the $i$-th to the $j$-th word in the $k$-th document.
Pipeline-based methods~\cite{chen2017reading,lin2018denoising,wang2019multi} typically leverage a document retriever to reduce the number of documents to read, but they suffer from error propagation when wrong documents are retrieved and can be still slow due to the heavy reader model.

\paragraph{Phrase-Indexed Open-domain QA}
As an alternative, ~\citet{seo2018phrase,seo2019real} introduce an end-to-end, real-time open-domain QA approach to directly encode all phrases in documents agnostic of the question, and then perform similarity search on the encoded phrases. This is feasible by decomposing the scoring function $F$ into two functions, 
\begin{equation*}
\begin{split}
    \hat{a}
    &= \argmax_{\xm_{i:j}^k} H_{\xm}(\xm_{i:j}^k) \cdot H_{\qm}(\qm)
\end{split}
\end{equation*}
where $H_{\xm}$ is the query-agnostic phrase encoding, and $H_{\qm}$ is the question encoding, and $\cdot$ denotes a fast inner product operation.

\citet{seo2019real} propose to encode each phrase (and question) with the concatenation of a dense vector obtained via a deep contextualized word representation model~\citep{devlin2019bert} and a sparse vector obtained via computing the tf-idf of the document (paragraph) that the phrase belongs to.
We argue that the inherent characteristics of tf-idf, which is not learned and identical across the same document, has limited representational power.
Our goal in this paper is to propose a better and learned sparse representation model that can further improve the QA accuracy in the phrase retrieval setup.

\section{Sparse Encoding of Phrases}
Our sparse model, unlike pre-computed sparse embeddings such as tf-idf, \emph{dynamically} computes the weight of each n-gram that depends on the context.

\subsection{Why do we need sparse representations?}
To answer the question in Figure~\ref{fig:overview}, the model should know that the target answer (\textit{\textbf{415,000}}) corresponds to the year \textit{\underline{1991}} while the (confusing) phrase \textit{587,000} corresponds to the year \textit{2000}.
The dense phrase encoding is likely to have difficulty in precisely differentiating between \textit{\underline{1991}} and \textit{2000} since it needs to also encode several different kinds of information.
Window-based tf-idf would not help because the year \textit{2000} is closer (in word distance) to \textit{\textbf{415,000}}.
This example illustrates the strong need to create an n-gram-based sparse encoding that is highly syntax- and context-aware.

\subsection{Contextualized Sparse Representations}
The sparse representation of each phrase is obtained as the concatenation of its start word's and end word's sparse embedding, i.e. $\sbo_{i:j} = [\sbo^\text{start}_i, \sbo^\text{end}_j]$. This way, similarly to how the dense phrase embedding is obtained in \citet{seo2019real}, we can efficiently compute them without explicitly enumerating all possible phrases.

We obtain each (start/end) sparse embedding in the same way (with unshared parameters), so we just describe how we obtain the start sparse embedding here and omit the superscript `start'.
Given the contextualized encoding of each document $\Hb = [\hb_1,\dots,\hb_{N}] \in \mathbb{R}^{N \times d}$, 
we obtain its (start/end) sparse encoding $\Sb = [\sbo_1, \dots, \sbo_N] \in \mathbb{R}^{N \times F}$ by
\begin{equation}\label{eqn:sparse}
    \mathbf{S} = \text{ReLU}\big( \frac{\mathbf{Q}\mathbf{K}^\top}{\sqrt{d}} \big)\mathbf{F} \in \mathbb{R}^{N \times F}
\end{equation}
\noindent where $\mathbf{Q}, \mathbf{K} \in \mathbb{R}^{N \times d}$ are query, key matrices obtained by applying a (different) linear transformation on $\Hb$ (i.e., using $W_\mathbf{Q}, W_\mathbf{K}: \mathbb{R}^{N \times d} \rightarrow \mathbb{R}^{N \times d}$), and $\mathbf{F} \in \mathbb{R}^{N \times F}$ is an one-hot n-gram feature representation of the input document $\xm$.
That is, for instance, if we want to encode unigram (1-gram) features, $\mathbf{F}_i$ will be a one-hot representation of the word $x_i$, and $F$ will be equivalent to the vocabulary size. 
Intuitively, $\sbo_i$ contains a weighted bag-of-ngram representation where each n-gram is weighted by its relative importance on each start or end word of a phrase.
Note that $F$ will be very large, so it should always exist as an efficient sparse matrix format (e.g., csc), and one should not explicitly create its dense form.
Since we want to handle several different sizes of n-grams, we create the sparse encoding $\Sb$ for each n-gram and concatenate the resulting sparse encodings.
In practice, we experimentally find that unigram and bigram are sufficient for most use cases.

We compute sparse encodings on the question side ($\sbo' \in \mathbb{R}^{F}$) in a similar way to the document side, with the only difference that we use the \texttt{[CLS]} token instead of start and end words to represent the entire question.
We share the same BERT and linear transformation weights used for the phrase encoding.

\subsection{Training}\label{sec:training}
As training phrase encoders on the whole Wikipedia is computationally prohibitive, we use training examples from an extractive question answering dataset (SQuAD) to train our encoders.
We also use an improved negative sampling method which makes both dense and sparse representations more robust to noisy texts.

\paragraph{Kernel Function}
Given a pair of question $\qm$ and a golden document $\xm$ (a paragraph in the case of SQuAD), we first compute the dense logit of each phrase $\xm_{i:j}$ by $l_{i,j} = \hb_{i:j} \cdot \hb'$.
Each phrase's sparse embedding is trained, so it needs to be considered in the loss function.
We define the sparse logit for phrase $\xm_{i:j}$ as $l_{i,j}^\text{sparse} = \mathbf{s}_{i:j} \cdot \mathbf{s}_\texttt{[CLS]}' = \mathbf{s}_i^\text{start} \cdot {\mathbf{s}'}_\texttt{[CLS]}^\text{start} + \mathbf{s}_j^\text{end} \cdot {\mathbf{s}'}_\texttt{[CLS]}^\text{end}$.
For brevity, we describe how we compute the first term $\mathbf{s}_i^\text{start} \cdot {\mathbf{s}'}_\texttt{[CLS]}^\text{start}$ corresponding to the start word (and dropping the superscript `start'); the second term can be computed in the same way.

\begin{eqnarray}
\mathbf{s}_i^\text{start} \cdot {\mathbf{s}'}_\texttt{[CLS]}^\text{start} =& \\
\text{ReLU}\big( \frac{\mathbf{Q}\mathbf{K}^\top}{\sqrt{d}} \big)_i\mathbf{F}&\bigg(\text{ReLU}\big( \frac{\mathbf{Q}'{\mathbf{K}'}^\top}{\sqrt{d}} \big)_{\texttt{[CLS]}}\mathbf{F}'\bigg)^\top \notag
\end{eqnarray}
where $\mathbf{Q}', \mathbf{K}' \in \mathbb{R}^{M \times d}, \mathbf{F}' \in \mathbb{R}^{M \times F}$ denote the question side query, key, and n-gram feature matrices, respectively. The output size of $\mathbf{F}$ is prohibitively large, but we efficiently compute the loss by precomputing $\mathbf{F} {\mathbf{F}'}^\top \in \mathbb{R}^{N \times M}$.
Note that $\mathbf{F} {\mathbf{F}'}^\top$ can be considered as applying a kernel function, i.e. $K(\mathbf{F}, \mathbf{F}') = \mathbf{F} {\mathbf{F}'}^\top$ where its $(i,j)$-th entry is $1$ if and only if the n-gram at the $i$-th position of the context is equivalent to the $j$-th n-gram of the question, which can be efficiently computed as well. One can also think of this as \emph{kernel trick} (in the literature of SVM~\citep{cortes1995support}) that allows us to compute the loss without explicit mapping.

The final loss to minimize is computed from the negative log likelihood over the sum of the dense and sparse logits:
\begin{equation*}\label{eq:loss}
    L = -(l_{i^*,j^*}+l_{i^*,j^*}^\text{sparse}) + \log\sum_{i,j}\exp(l_{i,j} + l_{i,j}^\text{sparse})
\end{equation*}
where $i^*, j^*$ denote the true start and end positions of the answer phrase. 
As we don't want to sacrifice the quality of dense representations which is also very critical in dense-first search explained in Section~\ref{sec:imple_detail}, we add dense-only loss that omits the sparse logits (i.e. original loss in~\citet{seo2019real}) to the final loss, in which case we find that we obtain higher-quality dense phrase representations.

\paragraph{Negative Sampling}
To learn robust phrase representations, we concatenate negative paragraphs to the original SQuAD paragraphs.
To each paragraph $\xm$, we concatenate the paragraph $\xm_\text{neg}$ which was paired with the question whose dense representation $\hb'_\text{neg}$ is most similar to the original dense question representation $\hb'$, following ~\citet{seo2019real}. 
We find that adding tf-idf matching scores on the word-level logits of the negative paragraphs further improves the quality of sparse representations.

\section{Experiments}
\subsection{Experimental Setup}
\paragraph{Datasets}
\squadopen~is the open-domain version of SQuAD~\citep{rajpurkar2016squad}.
We use 87,599 examples with the golden evidence paragraph to train our encoders and use 10,570 examples from dev set to test our model, as suggested by~\citet{chen2017reading}.
\textsc{CuratedTrec} consists of question-answer pairs from TREC QA~\citep{voorhees1999trec} curated by~\citet{baudivs2015modeling}.
We use 694 test set QA pairs for testing our model.
We only train on SQuAD and test on both \squadopen~and \trec, relying on the generalization ability of our model (zero-shot) for CuratedTREC.

\paragraph{Implementation Details}\label{sec:imple_detail}
We use and finetune BERT-Large for our encoders.
We use BERT vocabulary which has 30,522 unique tokens based on byte pair encodings.
As a result, we have $F\approx 1B$ when using both uni-/bigram features.
We do not finetune the word embedding during training.
We pre-compute and store all encoded phrase representations of all documents in Wikipedia (more than 5 million documents).
It takes 600 GPU hours to index all phrases in Wikipedia. 
We use the same storage reduction and search techniques by \citet{seo2019real}.
For search, we perform dense search first and then rerank with sparse scores (DFS) or perform sparse search first and rerank with dense scores (SFS), or a combination of both (Hybrid).

\paragraph{Comparisons}
For models using dedicated search engines, we show performances of DrQA~\citep{chen2017reading}, R$^3$~\citep{wang2018r}, Paragraph Ranker~\citep{lee2018ranking}, Multi-Step-Reasoner~\citep{das2018multistep}, BERTserini~\cite{yang2019end}, and Multi-passage BERT~\citep{wang2019multi}.
For end-to-end models that do not rely on search engine results, \textsc{DenSPI}~\citep{seo2019real}, ORQA~\citep{lee2019latent}, and \textsc{DenSPI} + \ours~ (Ours) are evaluated.
For \textsc{DenSPI} and ours, `Hybrid' search strategy is used.

\subsection{Results}
\begin{table}[t]
\small
\centering
\resizebox{1.0\columnwidth}{!}{
\begin{threeparttable}
\def\arraystretch{1.2}
\begin{tabular}{lcccc}
\toprule
\multirow{2}{1cm}{\textbf{Model}} &\multicolumn{1}{c}{\textsc{C.TREC}} & \multicolumn{2}{c}{\squadopen} & \\ 
& \fontsize{8}{9.6} EM & \fontsize{8}{9.6} EM & \fontsize{8}{9.6} F1 & \fontsize{8}{9.6} s/Q \\
\midrule
\multicolumn{5}{l}{\textit{Models with Dedicated Search Engines}} \\
\midrule
DrQA & 25.4\tnote{*} & 29.8\tnote{**} & - & 35 \\
R$^3$ & 28.4\tnote{*} & 29.1 & 37.5 & - \\ 
Paragraph Ranker & \textbf{35.4}\tnote{*} & 30.2 & - & 161 \\
Multi-Step-Reasoner & - & 31.9 & 39.2 & - \\
BERTserini & - & 38.6 & 46.1 & 115 \\
Multi-passage BERT & - & \textbf{53.0} & \textbf{60.9} & 84 \\ \midrule
\multicolumn{5}{l}{\textit{End-to-End Models}} \\
\midrule
ORQA & 30.1 & 20.2 & - & 8.0 \\ 
\textsc{DenSPI} &
31.6\tnote{$\dagger$} & 36.2 & 44.4 & 0.71 \\
\textsc{DenSPI} + \ours~(Ours) & \textbf{35.7}\tnote{$\dagger$} & \textbf{40.7} & \textbf{49.0} & 0.78 \\
\bottomrule
\end{tabular}
\begin{tablenotes}\footnotesize
\item[*] Trained on distantly supervised training data.
\item[**] Trained on multiple datasets
\item[$\dagger$] No supervision using target training data.
\end{tablenotes}
\end{threeparttable}}
\caption{Results on two open-domain QA datasets. See Appendix~\ref{sec:appendix_complexity} for how s/Q is computed.
}\vspace{-.3cm}\label{tab:odqa}
\end{table}

\paragraph{Open-Domain QA Experiments} Table~\ref{tab:odqa} shows experimental results on two open-domain question answering datasets, comparing our method with previous pipeline and end-to-end approaches. 
On both datasets, our model with contextualized sparse representations (\textsc{DenSPI} + \ours) largely improves the performance of the phrase-indexing baseline model (\textsc{DenSPI}) by more than 4\%.
Also, our method runs significantly faster than other models that need to run heavy QA models during the inference.
On \trec, which is constructed from real user queries, our model achieves state-of-the-art performance at the time of submission.
Even though our model is only trained on SQuAD (i.e., zero-shot), it outperforms all other models which are either distant- or semi-supervised with at least \fasterleast x faster inference.

On \squadopen, our model outperforms BERT-based pipeline approaches such as BERTserini~\citep{yang2019end} while being more than two orders of magnitude faster.
Multi-passage BERT, which utilizes a dedicated document retriever, outperforms all end-to-end models with a large margin in \squadopen.
While our main contribution is on the improvement in end-to-end, we also note that retrieving correct documents in \squadopen~is known to be often easily exploitable~\cite{lee2019latent}, so we should use more open-domain-appropriate test datasets (such as \trec) for a more fair comparison.

\begin{table}
\center
\resizebox{.9\columnwidth}{!}{%
\begin{tabular}{lcc}
\toprule
\multirow{1}{1cm}{\textbf{Model}} & \multicolumn{1}{c}{SQuAD$_\textsc{1/100}$} & \multicolumn{1}{c}{SQuAD$_\textsc{1/10}$} \\
\midrule
Ours & 60.0 & 51.6 \\
$-$ \ours & 55.9 ($-$4.1) & 48.4 ($-$3.2)\\
$-$ Doc./Para. tf-idf & 58.1 ($-$1.9) & 50.9 ($-$0.7) \\
$+$ Trigram~\ours & 58.0 ($-$2.0) & 49.8 ($-$1.8) \\
\bottomrule
\\
\end{tabular}}

\vspace{-.3cm}
\caption{Ablations of our model. We show effects of different sparse representations.}\label{tab:ablation}
\end{table}

\begin{table}[]
    \centering
    \resizebox{.95\columnwidth}{!}{%
    \begin{tabular}{p{1.5cm}lcc}
        \toprule
                      & \textbf{Model}   & EM    & F1 \\
        \midrule
         
        \multirow{2}{2cm}{Original} &  DrQA~\citep{chen2017reading}  & 69.5 & 78.8 \\
        & BERT~\citep{devlin2019bert} & 84.1 & 90.9 \\
        \midrule
         
        \multirow{3}{2cm}{Query-Agnostic} & LSTM + SA + ELMo      & 52.7 & 62.7\\
        & \textsc{DenSPI}      & 73.6 & 81.7\\
        & \textsc{DenSPI} +~\ours      & 76.4 & 84.8\\
        \bottomrule
    \end{tabular} 
    }
    \caption{Results on the SQuAD development set. LSTM+SA+ELMo is a query-agnostic baseline from~\citet{seo2018phrase}.
    \label{tab:pi-squad}}\vspace{-.3cm}
\end{table}

\begin{table*}[t]
    \centering
    \resizebox{1.9\columnwidth}{!}{
    \begin{tabular}{ll}
        \toprule
        \multicolumn{2}{l}{Q: When is Independence Day?} \\
        \midrule
        DrQA & [\textit{Independence Day (1996 film)}] Independence Day is a \textbf{1996} American science fiction ... \\
        \textsc{DenSPI}~+~\ours & [\textit{Independence Day (India)}] ... is annually observed on \textbf{15 August} as a national holiday in India. \\
        \midrule
        
        \multicolumn{2}{l}{Q: What was the GDP of South Korea in 1950?}\\
        \midrule
        \textsc{DrQA} & [\textit{Economy of South Korea}] In 1980, the South Korean GDP per capita was \textbf{\$2,300}. \\
        \textsc{DenSPI} & [\textit{Economy of South Korea}] In 1980, the South Korean GDP per capita was \textbf{\$2,300}. \\
        \textsc{DenSPI}~+~\ours & [\textit{Developmental State}] South Korea's GDP grew from \textbf{\$876} in 1950 to \$22,151 in 2010.  \\
        \bottomrule
        
    \end{tabular}
    }
    \caption{Prediction samples from DrQA, \textsc{DenSPI}, and \textsc{DenSPI}~+~\ours. Each sample shows [\textit{document title}], context, and \textbf{predicted answer}.}
    \label{tab:sparse_samples}\vspace{-.3cm}
\end{table*}




\paragraph{Ablation Study} Table~\ref{tab:ablation} shows the effect of contextualized sparse representations by comparing different variants of our method on \squadopen. We use a subset of Wikipedia dump (1/100 and 1/10).
Interestingly, adding trigram features in \ours~is worse than using uni-/bigram representations only, calling for a stronger regularization for high-order n-gram features.
See Appendix~\ref{sec:appendix_strat} on how \ours~performs in different search strategies.

\paragraph{Closing the Decomposability Gap}
Table~\ref{tab:pi-squad} shows the performance of DenSPI +~\ours~in the SQuAD v1.1 development set, where a single paragraph that contains an answer is provided in each sample.
While BERT-Large that jointly encodes a passage and a question still has a higher performance than ours, we have closed the gap to 6.1 F1 score in a query-agnostic setting.

\paragraph{Qualitative Analysis} Table~\ref{tab:sparse_samples} shows the outputs of three OpenQA models: DrQA~\citep{chen2017reading}, \textsc{DenSPI}~\citep{seo2019real}, and \textsc{DenSPI} +~\ours~(ours).
Our model is able to retrieve various correct answers from different documents, and it often correctly answers questions with specific dates or numbers compared to \textsc{DenSPI} showing the effectiveness of learned sparse representations.

\section{Conclusion}
In this paper, we demonstrate the effectiveness of contextualized sparse representations, \ours, for encoding phrase with rich lexical information in open-domain question answering.
We efficiently train our sparse representations by kernelizing the sparse inner product space.
Experimental results show that our fast open-domain QA model that augments \textsc{DenSPI} with \ours~outperforms previous open-domain QA models, including recent BERT-based pipeline models, with two orders of magnitude faster inference time.

\section*{Acknowledgments}
This research was supported by National Research Foundation of Korea (NRF-2017R1A2A1A17069\\645, NRF-2017M3C4A7065887), ONR N00014-18-1-2826, DARPA N66001-19-2-403, Allen Distinguished Investigator Award, and Sloan Fellowship.
We thank the members of Korea University, University of Washington, NAVER Clova AI, and the anonymous reviewers for their insightful comments.

\bibliography{anthology,acl2020}
\bibliographystyle{acl_natbib}

\clearpage
\appendix
\setcounter{page}{1}

\section{Inference Speed Benchmark of Open-Domain QA Models}\label{sec:appendix_complexity}

Table~\ref{tab:complexity} shows how the inference speed of each open-domain QA model is estimated in our benchmarks. Many of these models are not open-sourced but based on BERT, so we use the speed of BERT on the given length token as the basis for computing the inference speed.

We also note that our reported number for the inference speed of DenSPI~\cite{seo2019real} is slightly faster than that reported in the original paper.
This is mostly because we are using a PCIe-based SSD (NVMe) instead of SATA-based.
We also expect that the speed-up can be greater with Intel Optane which has faster random access time.


\section{Model Performances in Different Search Strategies}\label{sec:appendix_strat}

\begin{table}[ht]
\center
\def\arraystretch{1.0}
\resizebox{1.0\columnwidth}{!}{%
\begin{tabular}{lcccc}
\toprule
\multirow{2}{1cm}{\textbf{Model}} &\multicolumn{2}{c}{\squadopen} & \multicolumn{2}{c}{\trec} \\
 & \multicolumn{1}{c}{\textsc{DenSPI}} & \multicolumn{1}{c}{+ \ours} & \multicolumn{1}{c}{\textsc{DenSPI}} & \multicolumn{1}{c}{+ \ours} \\
\midrule
\textsc{SFS} & 33.3 & 36.9 ($+$3.6) & 28.8 & 30.0 ($+$1.2) \\
\textsc{DFS} & 28.5 & 34.4 ($+$5.9) & 29.5 & 34.3 ($+$4.8) \\
\textsc{Hybrid} & 36.2 & 40.7 ($+$4.5) & 31.6 & 35.7 ($+$4.1) \\
\bottomrule
\end{tabular}}
\caption{Exact match scores of~\ours~in different search strategies. \textsc{SFS}: Sparse First Search. \textsc{DFS}: Dense First Search. \textsc{Hybrid}: Combination of \textsc{SFS} + \textsc{DFS}. Exact match scores are reported.}\label{tab:strat}
\end{table}

In Table~\ref{tab:strat}, we show how we consistently improve over \textsc{DenSPI} when \ours~is added in different search strategies.
Note that on~\trec~where the questions more resemble real user queries, \textsc{DFS} outperforms \textsc{SFS} showing the effectiveness of dense search when not knowing which documents to read.

\begin{table*}[ht]
\small
\centering
\resizebox{2.0\columnwidth}{!}{
\begin{threeparttable}
\def\arraystretch{1.0}

\begin{tabular}{lcccc}
\multicolumn{5}{l}{\textit{Models using Bi-LSTM as Base Encoders}} \\
\midrule
\multicolumn{1}{l}{Model}&\multicolumn{1}{c}{\# of Docs. to Read} & \multicolumn{1}{c}{\# of Docs. to Re-rank} & & \multicolumn{1}{c}{s/Q} \\
\midrule
DrQA & 5 & None & & 35 \\
Paragraph Ranker & $\approx 3$ & 20 & & 161 \\ \\

\multicolumn{5}{l}{\textit{Models using BERT as Base Encoders}} \\ 
\midrule
\multicolumn{1}{l}{Model}&\multicolumn{1}{c}{\# of Docs. to Read} & \multicolumn{1}{c}{\# of Docs. to Re-rank} & \multicolumn{1}{c}{Max. Sequence Length/Doc. Stride} & s/Q\\
\midrule
BertSerini & 100 paragraphs\tnote{*} & 100 paragraphs\tnote{*} & 384/128 & 115 \\
Multi-Passage BERT & 30 passages\tnote{**} & 100 passages\tnote{**} & 100/50 & 84 \\ 
ORQA & $k$ blocks\tnote{$\dagger$} & None & 384/128 & 8 \\ 
\end{tabular}
\begin{tablenotes} \footnotesize
\item[*] Assumed 1 paragraph = 200 BERT tokens.
\item[**] 1 passage = 100 BERT tokens.
\item[$\dagger$] 1 block = 288 BERT tokens. Assumed $k=10$.
\end{tablenotes}
\end{threeparttable}}
\caption{Details on how the inference speed of each open-domain QA model is estimated using a single CPU. We use the default 384/128 (Max. Sequence Length/Doc. Stride) setting for both BertSerini and ORQA.}\label{tab:complexity}
\end{table*}

\end{document}